\documentclass[journal]{IEEEtran}
\usepackage{bm}
\newcommand{\TrainingSet}{\bm{T}} 
\newcommand{\ValidationSet}{\bm{V}} 
\newcommand{\TestSet}{\Psi}
\newcommand{\MeanSubsample}{{\mu}}

\newcommand{\CovarianceMatrix}{\bm{C}}
\newcommand{\MahalanobisDistance}{\bm{D}}
\newcommand{\Sample}{\bm{x}}
\newcommand{\ContaminationRate}{\lambda}
\newcommand{\NumberOfKernels}{n}
\newcommand{\NumberOfClasses}{C}
\newcommand{\NumberOfBands}{b}

\newcommand{\KernelSize}{k}

\usepackage{amsfonts}
\usepackage{graphicx}
\usepackage{makecell}
\usepackage{tikz}
\usepackage{collcell}

\usepackage{colortbl}
\usepackage{pgfplots}
\usepackage{pgfplotstable}
\usepackage{multirow}
\usepackage{url}
\usepackage{rotating}
\usepackage{enumitem}
\usepackage[noadjust]{cite}

\usepackage{tikz, rotating, makecell}
\usetikzlibrary{arrows, positioning, decorations.pathreplacing}
\usepgflibrary{shapes.multipart}

\usepackage{pgfplots}
\usepgfplotslibrary{groupplots}
\pgfplotsset{major grid style={densely dotted, draw=black!100,  line width=.5pt}}

\ifCLASSINFOpdf
\else
\fi
\newcommand{\Activation}{$\mathit{Z}^ {l}$}
\newcommand{\SoftmaxActivation}{$\hat{\mathit{Z}^{l}}$}

\newcommand{\ConfidenceScore}{$c$}

\pgfplotstableset{
    /color cells/min/.initial=0,
    /color cells/max/.initial=1000,
    /color cells/textcolor/.initial=,
    color cells/.code={%
        \pgfqkeys{/color cells}{#1}%
        \pgfkeysalso{%
            postproc cell content/.code={%
                \begingroup
                \pgfkeysgetvalue{/pgfplots/table/@preprocessed cell content}\value
\ifx\value\empty
\endgroup
\else
                \pgfmathfloatparsenumber{\value}%
                \pgfmathfloattofixed{\pgfmathresult}%
                \let\value=\pgfmathresult
                \pgfplotscolormapaccess
                    [\pgfkeysvalueof{/color cells/min}:\pgfkeysvalueof{/color cells/max}]%
                    {\value}%
                    {\pgfkeysvalueof{/pgfplots/colormap name}}%
                \pgfkeysgetvalue{/pgfplots/table/@cell content}\typesetvalue
                \pgfkeysgetvalue{/color cells/textcolor}\textcolorvalue
                \toks0=\expandafter{\typesetvalue}%
                \xdef\temp{%
                    \noexpand\pgfkeysalso{%
                        @cell content={%
                            \noexpand\cellcolor[rgb]{\pgfmathresult}%
                            \noexpand\definecolor{mapped color}{rgb}{\pgfmathresult}%
                            \ifx\textcolorvalue\empty
                            \else
                                \noexpand\color{\textcolorvalue}%
                            \fi
                            \the\toks0 %
                        }%
                    }%
                }%
                \endgroup
                \temp
\fi
            }%
        }%
    }
}

\begin{document}

\title{Hyperspectral Band Selection Using Attention-based Convolutional Neural Networks}

\author{Pablo~Ribalta,~\IEEEmembership{Student Member,~IEEE,}
        Lukasz~Tulczyjew,
        Michal~Marcinkiewicz,
        and~Jakub Nalepa*,~\IEEEmembership{Member,~IEEE}
\thanks{This work was funded by European Space Agency (HYPERNET project).}
\thanks{P.~Ribalta, L.~Tulczyjew, and J.~Nalepa are with Silesian University of Technology, Gliwice, Poland. L.~Tulczyjew, M.~Marcinkiewicz, and J.~Nalepa are with KP Labs, Gliwice, Poland.}
\thanks{*~Corresponding author: jnalepa@ieee.org}
}

\markboth{Submitted to IEEE Access}%
{Shell \MakeLowercase{\textit{et al.}}: Bare Demo of IEEEtran.cls for IEEE Journals}
\maketitle

\begin{abstract}
Hyperspectral imaging has become a mature technology which brings exciting possibilities in many domains, including satellite image analysis. However, the high dimensionality and volume of such imagery is a serious problem which needs to be faced in Earth Observation applications, where efficient acquisition, transfer and storage of hyperspectral images are key factors. To reduce the cost of transferring hyperspectral data from a satellite back to Earth, various band selection techniques have been proposed. In this letter, we introduce a new hyperspectral band selection algorithm which couples new attention-based convolutional neural networks with anomaly detection. Our rigorous experiments showed that the deep models equipped with the attention mechanism deliver high-quality classification, and consistently identify significant bands in the training data, permitting the creation of refined and extremely compact sets that retain the most meaningful features.
\end{abstract}


\IEEEpeerreviewmaketitle

\section{Introduction}

Current advancements in the sensor technology bring exciting possibilities in hyperspectral satellite imaging (HSI) which is being actively applied in various domains, including precision agriculture, surveillance, military, land cover applications, and more~\cite{Dundar2018}. It captures a wide spectrum of light for each pixel---such detailed information can be effectively exploited in HSI \emph{classification} (assigning a class label to each pixel) and \emph{segmentation} (determining the boundaries of objects of a given class in an input HSI)~\cite{8642388}. However, hyperspectral data's high dimensionality is an important challenge towards its efficient analysis, transfer, and storage. There are two approaches for dealing with such noisy, almost always imbalanced, and often redundant data. \emph{Feature extraction} algorithms (with principal component analysis and its variations being the mainstream~\cite{s18061978}) generate new low-dimensional descriptors from hyperspectral images (HSI), whereas \emph{feature selection} techniques retrieve a subset of all HSI bands carrying the most important information. Although the former approaches can be applied to reduced HSI sets, they are generally exploited to process raw HSI data, they are computationally-expensive, can suffer from band noisiness, and may not be interpretable~\cite{YANG2017396}. Band selection techniques are divided into \emph{filter} (unsupervised) and \emph{wrapper} (supervised) algorithms. Applied before classification, filter approaches do not require ground-truth data to select specific bands~\cite{803411,7214263,4378560}. They, however, suffer from several drawbacks: (i)~it is difficult to select the optimal dimensionality of the reduced feature space, (ii)~band correlations are often disregarded, leading to the data redundancy~\cite{YANG2017396}, (iii)~bands which might be informative when combined with others are removed, and (iv)~noisy bands are often labeled as informative due to low correlation with other bands. Wrapper approaches use the classifier performance as the objective function for optimizing the subset of HSI bands~\cite{7378270,7817787}. Although these methods alleviate the computational burden of the HSI analysis, such algorithms induce serious computational overhead. In this work, we mitigate this problem, and incorporate the selection process into the deep network training. Such approaches have not been explored in the literature so far.

Deep learning (DL) has enabled unprecedented achievements and established the state of the art in a plethora of domains, including HSI analysis~\cite{Bengio2016}. In general, the HSI segmentation algorithms encompass \emph{conventional} machine learning techniques which require feature engineering~\cite{Li2018}, and DL approaches~\cite{Gao_2018}. DL can conveniently elaborate \emph{spectral} features~\cite{Zhong2017} or both \emph{spectral} and \emph{spatial} features without any user intervention. These features are intrinsically extracted by the deep nets operating on the full HSI. Therefore, we need to face the aforementioned challenges concerning the high HSI dimensionality in both conventional and DL-powered segmentation approaches. Attention mechanisms allow humans and animals to effectively process enormous amount of visual stimuli by focusing only on the most-informative chunks of data. An analogous approach can be applied in DL to localize the most informative parts of an input image to \emph{focus} on. We build upon the \textit{painless attention} mechanism which is trained during the network's forward-backward pass~\cite{painlessattention}, and exploit it in our convolutional architectures for HSI band selection. To the best of our knowledge, attention mechanisms have been used neither for this purpose, nor for HSI segmentation before.

In this letter, we introduce a new HSI band selection method (Section~\ref{sec:method}) which exploits attention-based convolutional neural networks (CNNs). The goal of this system is to learn which bands convey the most important information, as an outcome of the training process, alongside a ready-to-use deep model. Thus, our method is an \emph{embedded} approach---the generation of attention heatmaps is embedded into the CNN training. These heatmaps quantify the \emph{importance} of specific parts of the spectrum, and they are later processed using an anomaly detection algorithm. We build upon our observation that only a (very) small subset of all bands within an original HSI convey the important information, and these bands can be seen as \emph{outliers} (the other bands, which are in the majority, are not informative). The contribution of this work is multi-fold:

\begin{itemize}[leftmargin=*]
\item[-] We introduce a new HSI band selection algorithm (Section~\ref{sec:method}) which couples attention-based CNNs and anomaly detection (Section~\ref{sec:anomaly_detection}) to find the most important bands.
\item[-] We introduce attention-based CNNs to extract attention heatmaps that show which parts of the spectrum are \emph{important} during the training. Our CNNs are \emph{spectral}, and use the spectral information during the classification. However, they could be potentially extended by incorporating the convolutional layers which would operate in the spatial dimension, as the attention modules are topology-agnostic.
\item[-] We performed a rigorous experimental study (Section~\ref{sec:experiments}) to: (i)~compare our technique with the state of the art in HSI band selection, (ii)~verify the impact of band selection on various supervised learners, (iii)~understand the impact of appending the attention modules to our CNNs, and (iv)~verify the statistical importance of the results.
\end{itemize}

\section{Method}\label{sec:method}

\subsubsection{General overview of the deep network architecture}

\begin{figure}[ht!]
\centering
\includegraphics[width=\columnwidth]{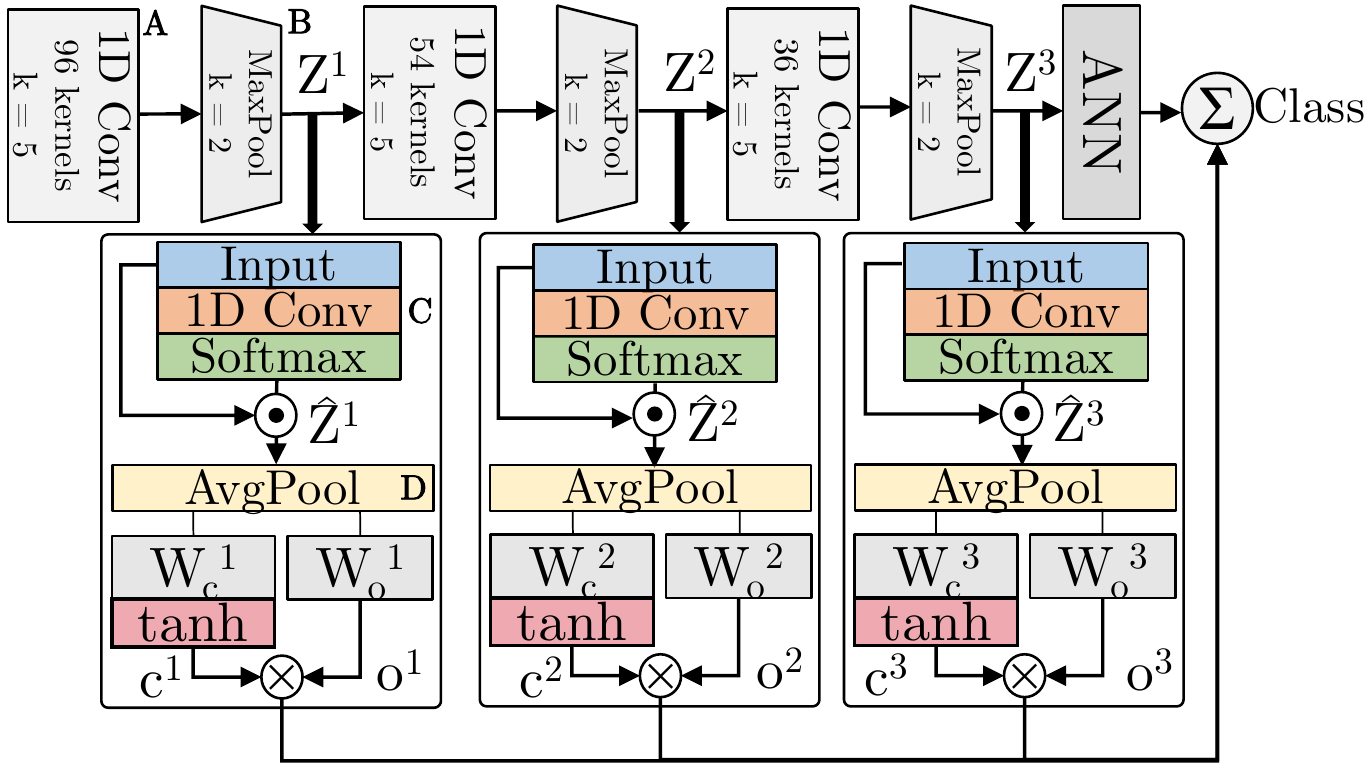}
\caption{In attention-based CNNs, features at different levels \Activation~are processed to generate the attention heatmaps, and they are used to output (i)~a class hypothesis based on the local information, and (ii)~a confidence score $c^{l}$. The final output is the softmaxed weighted sum of the attention estimators, and the output of the network's classifier (here, an artificial neural network, ANN containing two hidden layers containing 512 and 128 neurons with ReLU).}
\label{attention_mechanism}
\end{figure}

In the attention-based CNNs for HSI (Fig.~\ref{attention_mechanism}), an attention module is inserted after each {max-pooled} activation of a convolutional layer \Activation~($l$ denotes the depth within the network topology, and $l\geq 1$), in order to reduce the computational burden of the attention mechanism. This module is composed of two elements: an \emph{attention estimator}, extracting the most important regions of a feature map, and a \emph{confidence gate}, producing a confidence score for the prediction. We can easily modify the number of building blocks (BBs) in our CNNs---each BB encompasses the one-dimensional (1D) convolution followed by the non-linearity, batch normalization, and 1D max pooling layer (we operate only in the spectral dimension, hence both types of the layers are one-dimensional), alongside the attached attention module. We exploit the rectified linear unit (ReLU) as a non-linearity, which outputs zero for any negative input $x$, and it returns the value of $x$ otherwise. Hence, it can be formally written as ${\rm ReLU}(x)=\max\left(0, x\right)$. In this work, we experimentally analyzed the attention-based CNNs with two, three, and four BBs (Section~\ref{sec:experiments}).

Each layer in our CNN is parameterized with the corresponding hyper-parameter values: the number of kernels $\NumberOfKernels$, together with the size of the kernels in the convolutional layers, and the size of the pooling windows (both sizes are denoted as $\KernelSize$ in Fig.~\ref{attention_mechanism} for brevity). The kernel size is kept constant for all convolutional layers ($\KernelSize=5$, unit stride, and the padding of $2$). Similarly, the pooling window size was kept unchanged ($\KernelSize=2$ with the stride of $2$) in all max pooling layers. The number of kernels $\NumberOfKernels$ in the convolutional layers decreases with the increase of the number of BBs (Fig.~\ref{attention_mechanism}) to reduce the complexity of the model, alongside its memory footprint. We expect that the shallower convolutional layers will be able to extract low-level features based on the spectral information, as they should be manifested in specific parts of the spectrum.

In Table~\ref{tab:steps_dimensionality}, we present the dimensionality of the input and output tensors for the selected operations in our deep architecture (annotated as A, B, C, and D in Fig.~\ref{attention_mechanism}; $\NumberOfBands$ is the number of bands in the input HSI). The dimensions of the corresponding steps in the deeper parts of the CNN can be calculated analogously. The details of the C and D steps (in the attention module) are discussed in the following sections.

\begin{table}[ht!]
	\scriptsize
	\centering
	\caption{Data input and output dimensionality of selected steps in our attention-based CNNs (annotated as A, B, C, and D in Fig.~\ref{attention_mechanism}).}
	\label{tab:steps_dimensionality}
	\renewcommand{\tabcolsep}{0.4cm}
	\begin{tabular}{rll}
		\Xhline{2\arrayrulewidth}
		Step & Input & Output \\
		\hline
		A & $\NumberOfBands \times 1$ & $\NumberOfBands \times 96$\\
        B & $\NumberOfBands \times 96$ & $\left(\lfloor(\NumberOfBands-2)/2\rfloor+1\right) \times 96$\\
        C & $\left(\lfloor(\NumberOfBands-2)/2\rfloor+1\right) \times 96$ & $\left(\lfloor(\NumberOfBands-2)/2\rfloor+1\right) \times 1$\\
        D & $\left(\lfloor(\NumberOfBands-2)/2\rfloor+1\right) \times 96$ & $\left(\lfloor(\NumberOfBands-2)/2\rfloor+1\right) \times 1$\\
		\Xhline{2\arrayrulewidth}
	\end{tabular}
\end{table}

\subsubsection{Attention estimator}\label{sec:attention_estimator}

The attention estimator module encompasses the 1D convolution with one kernel applied (therefore, $\NumberOfKernels=1$; in~\cite{painlessattention}, the number of kernels is greater). The kernel size is equal to the number of feature maps extracted by the corresponding BB. This kernel performs the dimensionality reduction (all feature maps are reduced to one), and it is followed by the ReLU activation and spatial, across all entries within the feature map, {softmax} to elaborate the final heatmap. We do not softmax separate confidence scores, as suggested in~\cite{painlessattention}, to decrease the computational burden.

The attention estimator learns the following embedding:

\begin{equation}
\mathcal{F}: \mathbb{R}^{\NumberOfBands\times n}\rightarrow \mathbb{R}^{\NumberOfBands\times 1},
\end{equation}

\noindent where $\NumberOfBands$ and $n$ denote the number of HSI bands and the number of feature maps, respectively. The attention estimator merges all feature maps (FMs) at depth $l$ into a single one. The estimator builds a heatmap \SoftmaxActivation---it is used to normalize each map in \Activation, which denotes the set of all activation maps at the level $l$, i.e.,~the attention heatmap is exploited to highlight the importance of each entry in each map. The hypothesis $H^{l}$ of the output space given its local information is produced:



\begin{equation}
H^{l}={\rm AvgPool}(\hat{Z}^{l}\odot Z^{l}),
\end{equation}

\noindent where the $\odot$ sign is the element-wise product. The number of activation maps \Activation~at a given level $l$ is variable (i.e.,~96 after the first BB, 54 after the second BB, and 36 after the third BB), and this normalization is executed to each of them---they are average-pooled to produce the hypothesis $H^{l}$. It is used by a linear classifier to predict the label of the input sample:

\begin{equation}
o^{l}=H^{l}W_{o}^{l}.
\label{eq:output_classifier}
\end{equation}

\subsubsection{Confidence gate}

The local features are very often not enough to output a high-quality class hypothesis. Thus, we couple each attention module with the network's output to predict the confidence score \ConfidenceScore~by the means of an inner product with the gate weight matrix $W_c$ (at the $l$-\textit{th} level):

\begin{equation}
c^{l}=\tanh(H^{l}W^{l}_c).
\end{equation}

\noindent The network's outputis the softmaxed weighted sum of the attention estimators and the output of the classifier $o^{net}$ multiplied by its confidence score $c^{net}$:

\begin{equation}
{\rm output} = {\rm softmax}(o^{net}\cdot c^{net} + \sum\limits_{l=1}^{\left|\left|BB\right|\right|} c^{l}\cdot o^{l}).
\end{equation}

\noindent The softmax function converts a real-valued score $x$ into a probability value $p$ in the multi-class classification. Thus, a vector of such scores $\bm{x}\in\mathbb{R}^\NumberOfClasses$ is converted into a vector of probabilities $\bm{p}\in\left[0,1\right]^\NumberOfClasses$, where $p_i$ is the probability of an input pixel HSI belonging to the $i$-\textit{th} (out of $\NumberOfClasses$) class:
\begin{equation}
  p_i=\frac{e^{x_i}}{\sum_{k=0}^{\NumberOfClasses-1}e^{x_k}}.
\end{equation}
\noindent The output vector $\bm{p}$ is the probability vector, therefore it is non-negative, and $\sum_{c=0}^{\NumberOfClasses-1}p_c=1$, and can be used to predict the class label for each HSI pixel. To better understand the influence of the dimensionality reduction in the deeper parts of our attention-based CNNs, see an illustrative attention-based CNN example in the Supplementary Material\footnote{We use SM to refer to entities included in the Supplementary Material.} (Fig. 1SM).

\subsection{Selection of HSI bands as anomaly detection}\label{sec:anomaly_detection}

We exploit an Elliptical Envelope (EE) algorithm to extract the most discriminative bands based on the final attention heatmap. Since the number of important bands should be low, they can be understood as an \emph{anomaly} in the input set (Fig.1SM and Fig.~\ref{fig:average_heatmaps}). In EE, the data is modeled as a Gaussian distribution with covariances between feature dimensions---here, the entries of the heatmap. The input tensor to EE is $\NumberOfBands\times 1$, and an ellipse which covers the majority of the data is determined. The samples which lay outside of this ellipse are \emph{anomalous}~\cite{Hoyle2015anomaly}. EE utilizes a fast minimum covariance determinant estimator~\cite{doi:10.1080/00401706.1999.10485670}, where the data is divided into non-overlapping sub-samples for which the mean ($\MeanSubsample$) and covariance matrix in each feature dimension ($\CovarianceMatrix$) are calculated. The Mahalanobis distance $\MahalanobisDistance$ for each sample $\Sample$ is:
\begin{equation}
\MahalanobisDistance=\sqrt{(\Sample-\MeanSubsample)^T\CovarianceMatrix^{-1}(\Sample-\MeanSubsample)},
\end{equation}
\noindent and the samples with the smallest values of $\MahalanobisDistance$ are retained. In EE, the fractional contamination rate ($\ContaminationRate$) defines how much data in the analyzed dataset should be selected as anomalies. These data samples (i.e.,~spectral bands) are selected as \emph{important} in our band selection technique---they are assigned significantly larger attention values in the heatmap.


\section{Experiments}\label{sec:experiments}

\subsection{Experimental setup}\label{sec:experimental_setup}

In all experiments, we perform Monte-Carlo cross-validation and randomly divide each HSI dataset (Section~\ref{sec:datasets}) 30 times into balanced (with under-sampling) training ($\TrainingSet$) and validation ($\ValidationSet$) sets, and the unseen test sets ($\TestSet$). These sets encompass $80\%$, $10\%$, and $10\%$ of all pixels in the HSI, respectively, and they never overlap---since we analyze only spectral segmentation, this training-validation-test division does not cause a training-test information leak~\cite{8642388}. The $\TrainingSet$ and $\ValidationSet$ sets are used during the CNN training, whereas $\TestSet$ is utilized to quantify the generalization of the trained models. We report per-class, average accuracy (AA), and the values of the Cohen's kappa: $\kappa=1-\frac{1-p_o}{1-p_e}$, where $p_o$ and $p_e$ are the observed and expected agreement (assigned vs. correct class label), respectively, and $-1\leq \kappa\leq 1$ (the higher, the better). All the measures are averaged across all 30 runs.

Our CNNs were coded in \texttt{Python 3.6} with \texttt{PyTorch 0.4}---they are available at \url{https://github.com/ESA-PhiLab/hypernet/tree/master/python_research/experiments/hsi_attention}, alongside our implementations of other state-of-the-art methods (Section~\ref{sec:comparison_SOTA}). The CNN training (ADAM~\cite{DBLP:journals/corr/KingmaB14}; learning rate of $0.001$, $\beta_1 = 0.9$, and $\beta_2 = 0.999$) terminates if after 25 epochs the accuracy over $\ValidationSet$ does not increase.


\subsection{Datasets}\label{sec:datasets}

We focused on two imbalanced multi-class HSI benchmarks: Salinas Valley (acquired using the NASA Airborne Visible/Infrared Imaging Spectrometer AVIRIS sensor), and Pavia University (Reflective Optics System Imaging Spectrometer ROSIS sensor). AVIRIS registers 224 contiguous bands with wavelengths in a 400 to 2450 nm range (visible to near-infrared), with 10 nm bandwidth, and it is calibrated to within 1 nm. ROSIS collects the spectral radiance data in 115 bands in a 430 to 850 nm range (4 nm nominal bandwidth).


\subsubsection{Salinas Valley}

This set ($217\times 512$ pixels) was captured over Salinas Valley, California, USA, with a spatial res. of 3.7 m. The image shows different sorts of vegetation (16 classes). The original data contains 224 bands, however 20 bands were removed by the authors of this set due to either atmospheric absorption or noise contamination (see \url{https://tiny.cc/grsl}).


\subsubsection{Pavia University}

This set ($340\times 610$ pixels) was captured over Pavia University, Italy, with a spatial res. of 1.3 m. It shows an urban scenery (9 classes) with 103 bands, as 12 most noisy bands (out of 115) were removed by its authors.

\begin{figure}[ht!]
	\centering
	\includegraphics[width=1\columnwidth]{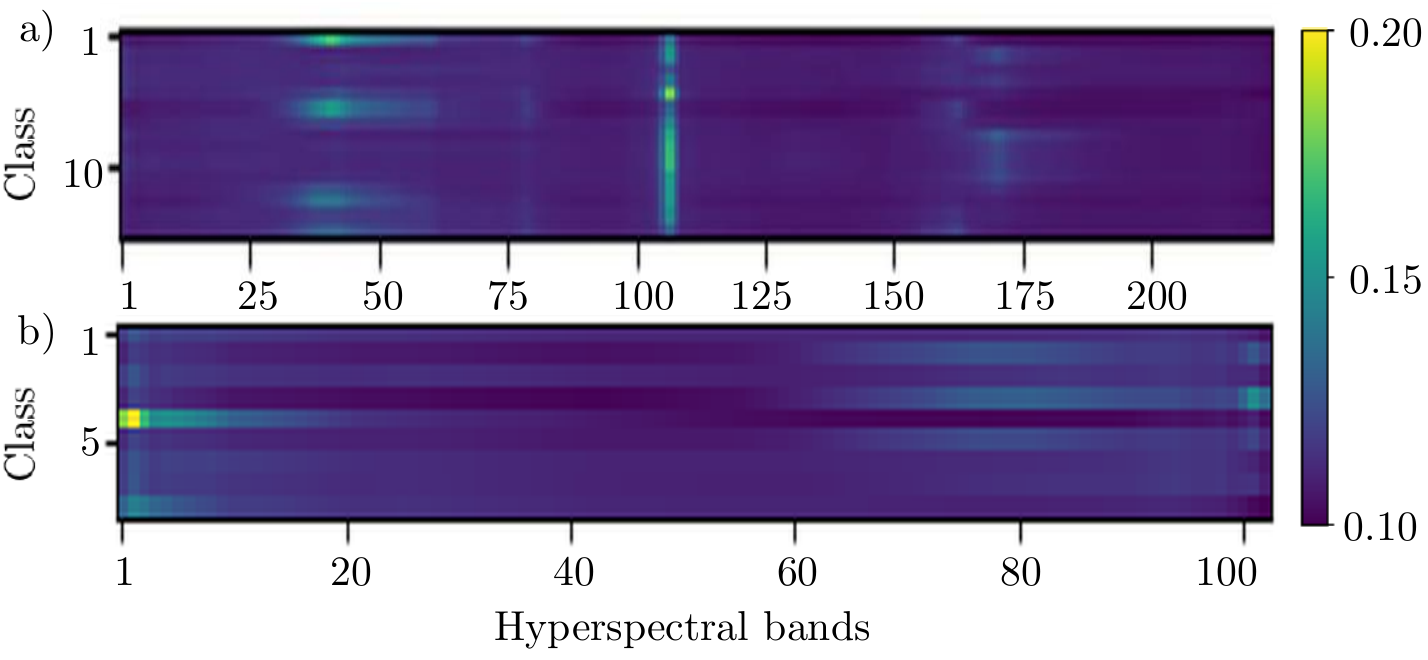}
	\caption{Example average attention-score heatmaps for a) Salinas Valley and b) Pavia University show that certain bands convey more information than the others (the brighter the regions are, the higher attention scores were obtained).}
	\label{fig:average_heatmaps}
\end{figure}

\subsection{Selection of bands using the proposed algorithm}\label{sec:selection_of_bands_using_attention_experiments}


We extracted bands from the benchmark HSI using our attention-based CNNs followed by the anomaly detection. For each set, we ran CNNs equipped with two, three, and four BBs (CNN-2A, CNN-3A, and CNN-4A) 30 times using Monte-Carlo cross-validation, and the attention scores (which were fairly consistent for all runs; $p<0.01$, two-tailed Wilcoxon tests) were averaged (see example in Fig.~\ref{fig:average_heatmaps}). Given the average attention scores, EE with different values of the contamination rate $\ContaminationRate=\{0.01, 0.02,\dots, 0.05\}$ was used to extract the final subset of HSI bands. The band selection results are gathered in Table~\ref{tab:number_of_selected_bands}. Although the contamination rate is a hyper-parameter of our method, the differences, in terms of the number of selected bands, across different $\ContaminationRate$ values are small. Our technique drastically decreased the number of HSI bands for all datasets, and for all $\ContaminationRate$'s (less than 14\% and 9\% of bands were selected as important for $\ContaminationRate=0.01$ for Salinas and Pavia, which amounts to 28 and only 9 bands, respectively).

\begin{table}[ht!]
	\scriptsize
	\centering
	\caption{Number of bands selected using the proposed algorithm for the a) Salinas Valley and b) Pavia University datasets.}
	\label{tab:number_of_selected_bands}
	\renewcommand{\tabcolsep}{0.23cm}
	\begin{tabular}{rrrrrrrr}
		\Xhline{2\arrayrulewidth}
		&Contamination rate ($\ContaminationRate$) $\rightarrow$ & 0.01 & 0.02 & 0.03 & 0.04 & 0.05 \\
		\hline
		\multirow{2}{*}{a)}   & Number of selected bands & 28 & 28 & 29 & 33 & 38 \\
		                      & Percentage of all bands & 13.73 & 13.73 & 14.22 & 16.18 & 18.63 \\
\hline
		\multirow{2}{*}{b)}   & Number of selected bands & 9 & 12 & 14 & 20 & 28 \\
		                      & Percentage of all bands & 8.74 & 11.65 & 13.59 & 19.42 & 27.18 \\
		\Xhline{2\arrayrulewidth}
	\end{tabular}
\end{table}

The average attention scores for Salinas and Pavia are visualized in Fig.~2SM. There are several attention peaks for Salinas Valley indicating the most meaningful part of the spectrum. Although for Pavia University there are less such clearly selected bands, some parts of the spectrum are definitely more distinctive than the others. This experiment showed that our method retrieves consistent attention scores annotating the most important bands, and that it is data-driven.

\subsection{Influence of attention modules on classification}

This experiment verifies whether applying attention modules in a CNN has any impact on its performance and convergence. For each set, we trained the deep networks with and without attention using original HSI data. The CNNs without attention are referred to as CNN-2, CNN-3, and CNN-4 (two, three, and four convolutional-pooling blocks, as depicted in Fig.~\ref{attention_mechanism}).

The average per-class accuracy scores (averaged across 30 executions) for Salinas and Pavia are gathered in Tables~1SM and 2SM. The differences between the architectures are not statistically important (i.e.,~CNN-2 compared with CNN-2A, CNN-3 with CNN-3A, and CNN-4 with CNN-4A), according to the Wilcoxon tests at $p<0.01$. Therefore, attention modules did not adversely impact the performance of the CNNs---they allow for building a high-quality model and selecting the most important bands \emph{at once}. Deeper CNNs delivered more stable results (std. dev. of the accuracy over $\TestSet$ decreased from 0.007 to 0.005 for Salinas, and from 0.03 to 0.01 for Pavia). On the other hand, we can observe only minor improvements in the performance when more BBs are appended. It shows that the shallower models can extract high-quality features using just two convolutional-pooling blocks. The same observation can be drawn from Figs.~3SM--4SM, where we render the kappa scores for Salinas and Pavia. There are classes (C8 and C15 for Salinas, and C1, C2 and C8 for Pavia) which are ``difficult'' for all classifiers (Tables~1SM--2M). In both cases, it is observed for the most numerous classes, and it can be attributed to the fact that they are under-sampled while creating the balanced training sets. Therefore, the sampled examples are not representative.

The average number of epochs before convergence, and the average processing time\footnote{Using NVIDIA Titan X Ultimate Pascal GPU 12 GB GDDR5X.} of a single epoch are presented in Figs.~5SM--6SM. Appending attention or adding BBs increases neither of them, hence they can be considered as a seamless CNN extension to enhance its operational ability.

\subsection{Classification accuracy over reduced datasets}\label{sec:influence_of_attention_on_accuracy}

We evaluated the performance of state-of-the-art models trained using full and reduced sets. They included Support Vector Machines (SVMs), Random Forests (RFs), and Decision Trees (DTs). We additionally executed grid search to optimize the hyper-parameters of all models: $C$ and $\gamma$ of the radial-basis kernel function in SVMs, number of trees in RFs, minimum samples per leaf in DTs, and minimum samples in a split in both RFs and DTs. The training with grid search was repeated 30 times (Monte-Carlo cross-validation). Table~3SM shows that decreasing HSI helps shorten the grid-search time which can easily become large for full sets. Hyper-parameter optimizations are not necessary in our CNNs.

The average-accuracy results gathered in Tables~1SM--2SM show that for most of the classes, the performance of the classifiers is not diminished by our band selection. Although there are classes for which the accuracy decreased (e.g., C2 and C3 in Pavia), the differences for other classes are negligible, especially for CNNs for $\ContaminationRate\geq0.03$. It is proved by the Wilcoxon tests executed to analyze the differences between models trained with different datasets (with and without reduction). Although the differences in AA of the classifiers trained with the reduced numbers of bands are statistically important (at $p<0.01$), they are not as dramatic as in other band selection algorithms~\cite{DBLP:journals/corr/abs-1802-06983}. Note that CNN-4A could not be trained for very small number of bands because of the dimensionality reduction in pooling layers.

The inference time of all investigated learners was very short. Reducing the number of bands decreased the \emph{total} inference time of all examples in $\TestSet$ which amounted to approx. 1,500 examples in Salinas, and to approx. 850 examples in Pavia for both sets: 0.06 down to 0.03~s (CNN-2A), 0.07 to 0.04~s (CNN-3A), 0.09 to 0.04~s (CNN-4A), and 0.16 to 0.12 (SVM) for Salinas Valley ($\ContaminationRate=0.01$; the time for RF and DT was unchanged and it amounted to 0.1~s and less than $0.01$~s). The decrease in the time was analogous for Pavia. Also, we do not report the times for CNNs without attention as they were practically the same as for the attention-based CNNs.

\subsection{Comparison with the state of the art}\label{sec:comparison_SOTA}

We compare our algorithm with other state-of-the-art techniques. For the sake of thoroughness, we took into consideration both filter and wrapper approaches. As a filter algorithm, we implemented the mutual information-based method (MI) \cite{MI}. In~\cite{MI}, the authors used the estimated reference maps to calculate the mutual information. Since this map should be estimated using available knowledge about the spectral signatures of the materials encountered within the scene, the lack of them may lead to incorrect maps. For fair comparison, we used the original ground-truth information instead of such estimated reference maps---it can render over-optimistic results for this method (i.e.,~our MI implementation is ``handicapped'' by the availability of ground truth). As the wrapper approaches, we selected two modern algorithms: a multi-objective immune algorithm (BOMBS)~\cite{BOMBS}, and the algorithm (ICM, \textbf{I}mproved \textbf{C}lassification \textbf{M}ap)~\cite{ICM}, in which the authors assess the quality of selected subsets of HSI bands using the pixel-wised classification map enhanced by the edge preserved filtering. We extracted the same number of bands as in Table~\ref{tab:number_of_selected_bands}.

The kappa scores are presented in Fig.7SM---for virtually all classifiers, they consistently grow for all techniques with the increase of the number of extracted bands, and ultimately converge to the same values. As already mentioned, the results for MI may be over-optimistic, as we utilize the entire ground-truth information to extract the important bands. Hence, we ``leak'' the information across the training and test sets because the training-validation-test splits are created \emph{after} the band selection step, and before training a supervised learner. The execution times (Table~4SM) show that our technique is orders of magnitude faster when compared with the wrapper algorithms (BOMBS and ICM) while delivering competitive classification results. Also, the number of bands selected in all methods was set according to our contamination factors---if we did not know the desired number of bands, we would have to execute each method in a grid search-like manner, and it would drastically increase their running time.


\section{Conclusion}\label{sec:conclusions}

We proposed new attention-based CNNs coupled with anomaly detection for selecting bands from HSI. The experimental validation showed that the proposed algorithm extracts important bands from HSI, and allows us to obtain state-of-the-art accuracy using only a fraction of bands (14--19\% for Salinas, and 9--27\% for Pavia). Overall, it revealed that:

\begin{itemize}[leftmargin=*]
\item[-] Attention-based CNNs deliver high-quality classification, and adding attention modules does not impact classification abilities and training time of an underlying CNN.
\item[-] Attention-based CNNs extract the most informative bands in HSI during the training in an embedded approach.
\item[-] Selected bands can be used to identify relevant and discard unimportant parts of the spectrum, drastically shortening training times of a classifier, and compressing HSI without sacrificing the amount of conveyed information.
\item[-] Our technique is applicable to any HSI set and any CNN.
\item[-] Our technique is competitive with the state-of-the-art approaches, and works orders of magnitude faster.
\end{itemize}

\ifCLASSOPTIONcaptionsoff
  \newpage
\fi

\bibliographystyle{ieeetran}
\bibliography{ref_all}

\end{document}